# An Efficient Intelligent System for the Classification of Electroencephalography (EEG) Brain Signals using Nuclear Features for Human Cognitive Tasks

Emad-ul-Haq Qazi[a], Muhammad Hussain[a] and Hatim Aboalsamh[a]

[a]*Visual Computing Lab, Department of Computer Science, College of Computer and Information Sciences, King Saud University, Riyadh 11543, Kingdom of Saudi Arabia.*

**Abstract.** Representation and classification of Electroencephalography (EEG) brain signals are critical processes for their analysis in cognitive tasks. Particularly, extraction of discriminative features from raw EEG signals, without any pre-processing, is a challenging task. Motivated by nuclear norm, we observed that there is a significant difference between the variances of EEG signals captured from the same brain region when a subject performs different tasks. This observation lead us to use singular value decomposition for computing dominant variances of EEG signals captured from a certain brain region while performing a certain task and use them as features (nuclear features). A simple and efficient class means based minimum distance classifier (CMMDC) is enough to predict brain states. This approach results in the feature space of significantly small dimension and gives equally good classification results on clean as well as raw data. We validated the effectiveness and robustness of the technique using four datasets of different tasks: fluid intelligence clean data (FICD), fluid intelligence raw data (FIRD), memory recall task (MRT), and eyes open / eyes closed task (EOEC). For each task, we analyzed EEG signals over six (06) different brain regions with 8, 16, 20, 18, 18 and 100 electrodes. The nuclear features from frontal brain region gave the 100% prediction accuracy. The discriminant analysis of the nuclear features has been conducted using intra-class and inter-class variations. Comparisons with the state-of-the-art techniques showed the superiority of the proposed system.

## 1. Introduction

EEG is a brain mapping technique to measure and assess the neurophysiological changes in human brain [1]. It enables the researchers to study brain processing for intelligence, vision, emotion, recognition, perception, motor imagery, and memory as well as to detect abnormalities like dementia, epilepsy, sleep disorders, stroke, trauma and depression. Existing techniques for the classification of EEG signals employ features extracted with statistical analysis, time–frequency analysis, spectral analysis, time series analysis and EEG rhythms analysis etc. [1]. However, these approaches do not give satisfactory results for the classification of EEG signals, especially when they are in their raw form, i.e., captured directly from brain and no pre-processing has been performed. It is because of the reason that EEG signals are highly vulnerable to artifacts due to their non-stationary characteristics [2]. It necessitates the need of a feature extraction technique that can extract discriminative features from raw EEG signals for their efficient and robust classification.

Nuclear norm has been shown to outperform $L_1$-norm, $L_2$-norm and Frobenius norm [3], especially in case of noisy data [4] and it is based on singular values (variances in the data). Motivated by nuclear norm, we propose nuclear features. The projection of raw EEG signals captured from a certain brain region on a singular space by singular value decomposition (SVD) gives the amount of variances along different directions. We observed that the small variances are due to the artifacts and dominant variances represent the discriminative part of the signals. In view of this, we propose to use dominant variances to represent an event. A small number of dominant variances is enough to discriminate two different events. As such the proposed feature extraction technique results in a feature space of small dimension, where the regions corresponding to different classes are well-separated. As such a simple class means based minimum distance classifier (CMMDC), which needs only class means, is enough to efficiently and reliable classify any unknown event.

The proposed approach has been validated on four different EEG datasets. A detailed evaluation of the proposed method indicated that it gives outstanding result for clean as well raw data. An overview of the overall system is given in Fig. 1

The main contributions are:

(i) A technique to extract nuclear features from EEG signals using SVD, which is different from other feature extraction techniques based on SVD in the sense that it uses the singular values (variances) as nuclear features that are effective in discriminating EEG signals corresponding to different events,

(ii) The nuclear features, which are robust and computationally efficient in representing raw EEG signals, save from laborious and time consuming process of cleaning EEG signals.

(iii) A simple, robust and memory-efficient classification technique, which does not involve learning and needs only class means.

(iv) A robust and efficient system for the classification of EEG brain signals, which represent two types of events; it is suitable for real time applications because of its efficiency.

(v) The discriminant analysis of the features using a well-known intra-class and inter-class variation analysis.

The rest of the paper is organized as follows: In Section 2, we present the literature review. Section 3 describes in detail the materials and methods. Experimental results and discussion are given in Section 4, while Section 5 concludes the paper.

## 2. Literature Review

In the literature, various EEG feature extraction methods have been reported, i.e., transform based approaches, spectral analysis, wavelet analysis, power analysis, entropy analysis, time–frequency analysis, and time series analysis [5-12]. In these methods, discriminative features are extracted from EEG data and given to different classifiers to classify EEG brain signals. We present a review of literature on feature extraction and classification of EEG brain signals.

Transform based technique is one of the important approaches to extract discriminative features from EEG signals. Its objective is to get lower dimensional information in compact form where maximum data energy is presented in a few coefficients which are uncorrelated. After removing the indiscriminative features, these techniques help to extract appropriate features which give better generalization performance in the classification by reducing the computational complexity [13]. For feature extraction and classification, EEG time series (entropy) analysis has been used to detect epileptic seizure and classification of control and schizophrenic subjects [7, 14-17]. These research studies extracted different features such as permutation entropy, sample entropy (SamEn) and approximate entropy (ApEn) for the classification of EEG signals. Spectral analysis of EEG signals has also been extensively used to extract features [18-20]. It comprises of EEG signals rhythms analysis like alpha, beta, theta, gamma and delta frequencies, power density spectrum, local minima and maxima, autoregressive moving average for the classification problem of EEG brain signals. The time-frequency analysis of EEG signals has been used for clinical EEG data to extract wavelet features from EEG patterns such as epileptic seizure detection [21-23]. Acharya et al. [7] extracted various features from EEG signals such as wavelet-based features, fractal dimension (FD), SamEn and ApEn. After feature extraction, the author used different classifiers like neural network (NN), k-nearest neighbor (k-NN), support vector machines (SVM) and decision tree to identify epileptic seizures. The study reported the accuracy of 99% by using wavelet and time-domain based features. However, the author used the small dataset of epilepsy, i.e., Bonn epilepsy dataset [24] as compared to large datasets such as Freiburg and CHBMIT dataset. Sabeti et al. [17] employed different features based on complexity and entropy such as FD, spectral entropy, ApEn and Lempel–Ziv complexity. The author performed the classification of EEG signals of schizophrenic patients and achieved the accuracy of 80-90 % by using Adaboost and linear discriminant analysis (LDA) classifiers. Taghizadeh-Sarabi et al. [25] extracted wavelet-based features from EEG signals by using three wavelets, i.e., Symlet2, Haar and Db4. SVM classifier has been used to classify various objects such as buildings, stationary, animals etc. through EEG signals. The author achieved the classification accuracy of 80% for stationary and animal objects groups. Zarjam et al. [26] and Wu and Neskovic [27] extracted wavelet complexity and entropy features from EEG signals and used non-linear classifiers like SVM and NN to classify working memory (WM) loads. They achieved the classification accuracy between 90-96% for discriminating various WM loads. However, the EEG brain signals used in these research studies are time-locked. Moreover, the length of EEG brain signals is relatively shorter than the spontaneous EEG brain signals. Jahidin et al. [28] obtained the accuracy of 88.89% by using power ratio of EEG sub-bands (alpha, beta and theta) as a feature. The author used the ANN as a classifier to classify EEG samples associated to various intelligent quotient classes.

The overview of the state-of-the-art methods given above indicates that most of the existing methods do not give satisfactory performance for classification of EEG signals. Moreover, the existing methods are application specific and work only on the pre-processed EEG signals [7, 26-27, 54]. Further, these techniques suffer from overfitting problem; when they are applied on different dataset concerning the same problem, the classification rate [29]. These drawbacks of the existing techniques motivated us to develop robust, efficient and effective method for the classification of EEG brain signals in their raw form, directly captured from brain and without applying any pre-processing.

## 3. Materials and Methods

To develop a system for robust and efficient classification of EEG signals, we collected EEG brain signals from thirty-four (34) subjects. For this purpose, 34 healthy male subjects were selected to participate in cognitive and baseline tasks. They were all healthy students. 31 were right-handed, and the remaining three were left handed students. Their age range was from 20 to 30 years. They were all medically fit and free from neurological disorders and hearing impairments, and were not using any medication. They possessed corrected to normal or normal vision. All subjects were briefed about the experiments. All of them showed their consent and signed the consent form before the test. The Human Research Ethics Committee of the Universiti Sains Malaysia and Ethics Coordination Committee of the Universiti Teknologi PETRONAS approved this research study.

In fluid intelligence clean data (FICD) and fluid intelligence raw data (FIRD), Raven's Advance Progressive Metric (RAPM) test was used and the subjects were divided into two (02) groups, i.e., high ability (HA) and low ability (LA) based on their intellectual ability. Next, we used the visual oddball cognitive task to capture the neural activity of each subject; in this task standard and target stimuli were presented to the subjects.

From the recorded EEG signals of fluid intelligence prediction level, we prepared two datasets for the classification of subjects into HA and LA groups based on their fluid intelligence level: $FIRD$, the raw dataset without any processing after recording the signal, $FICD$, the dataset made clean after recording by removing artifacts. The EEG trials of each subject were segmented utilizing a window size of duration six hundred (600) mSec, which comprises pre-stimulus period of one hundred (100) mSec (the baseline) and post-stimulus duration of five hundred (500) mSec.

For FICD, the data is cleaned by performing some processing operations. First, DC components and muscular artifacts associated to high frequency are removed utilizing a band pass filter (bpf), roll off 12 dB octave, 0.3-30Hz. After that, the trials which suffer from artifacts such as eye blinks and eye movements were removed, for example, EEG signal was rejected if its amplitude was ±90 μV. All recorded EEG signals were visual inspected and the channels. If they had no contact in widespread drift phase [30] then they were rejected. If any bad channel was found then spherical spline method [31] was utilized to remove a trial.

In memory recall task (MRT), the experiments, which were carried out comprised of two tasks, i.e., learning of educational contents and information recall from memory. In the learning task, the participants viewed 2D learning content for the time spans of 8 to 10 minutes. In the phase of memory recall, the retention period was of 30 minutes. In this process, twenty MCQs were inquired from the participants, and each MCQ had four choices for answers. While performing the recall experiment, EEG signals were recorded for study.

For MRT, raw EEG data was filtered by applying the band pass filter (1-48 Hz). The artifacts were identified. In next step, data was exported into .mat files format (Matlab) by utilizing the Netstation software of EGI. Visual examination, and Gratton and Coles method [32] were utilized to remove the ocular artifacts in the recorded data.

In eyes open / eyes closed (EOEC) dataset, thirty subjects (34) subjects were participated in the baseline task, i.e. EO and EC. EEG signals were recorded in each of the rest condition for the duration of 05 (five) minutes for each subject before memory recall task (MRT). For EOEC dataset, EEG recording and data preprocessing were performed similarly as discussed for MRT dataset.

The brain activation of each subject was captured as EEG signals from following different brain regions as shown in Fig. 2:

i. $TEMP$: Temporal-right (TR) and Temporal-left (TL) with 08 channels
ii. $FRONT$: Frontal-right (FR) and frontal-left (FL) with 16 channels
iii. $CENT$: Central-right (CR) and central-left (CL) with 20 channels
iv. $PERI$: Parietal-right (PR) and parietal-left (PL) with 18 channels
v. $OCCIP$: $Occipital - right$ (OR) and occipital-left (OL) with 18 channels
vi. $ALL$: All regions (AR) - 100 channels

Then nuclear features were extracted from raw and clean data using SVD based method. Finally, the discriminative features were input to the simple and efficient CMMDC to predict whether a subject belongs to HA or LA group in raw as well as clean data, classify the correct and incorrect answers, and classify EO and EC brain states. The detail of the experimental material, subjects, data collection procedure and processing EEG signals after recording is discussed in [33, 34].

### 3.1. Nuclear Features

In this section, we give the detail of the proposed method to extract discriminative features from EEG signals. The idea of feature extraction was motivated from nuclear norm [35], which is defined using singular values of a matrix as follows:

$$\|C\|_* = \sum_{i=1}^{n} \sigma_i(C), \quad (1)$$

where $\sigma_i$, $i = 1, 2, 3, \ldots, n$ are the singular values.

It has been shown that nuclear norm is more discriminative and robust than $L_1$-norm, $L_2$-norm and Frobenius norm [3] and has been employed for many pattern recognition tasks such as robust PCA [36], low rank matrix recovery [37, 38], nuclear norm based 2-DPCA (N-2-DPCA) [35]. The apparent reason that nuclear norm outperforms $L_1$-norm and $L_2$-norm is that it is based on singular values. This indicates that singular values can be considered to represent EEG brain signals. Based on this observation, we computed singular values of the nuclear matrices (defined later) of EEG signals measured from the same region (e.g. FRONT region – FL and FR in Fig. 2) corresponding to different brain states (events) and selected two dominant singular values to represent the EEG signals. The proposed feature extraction methodology is shown in Fig. 3. The plots of the singular values for different events are shown in Fig. 4 - 7. It further strengthened our idea to use singular values as features.

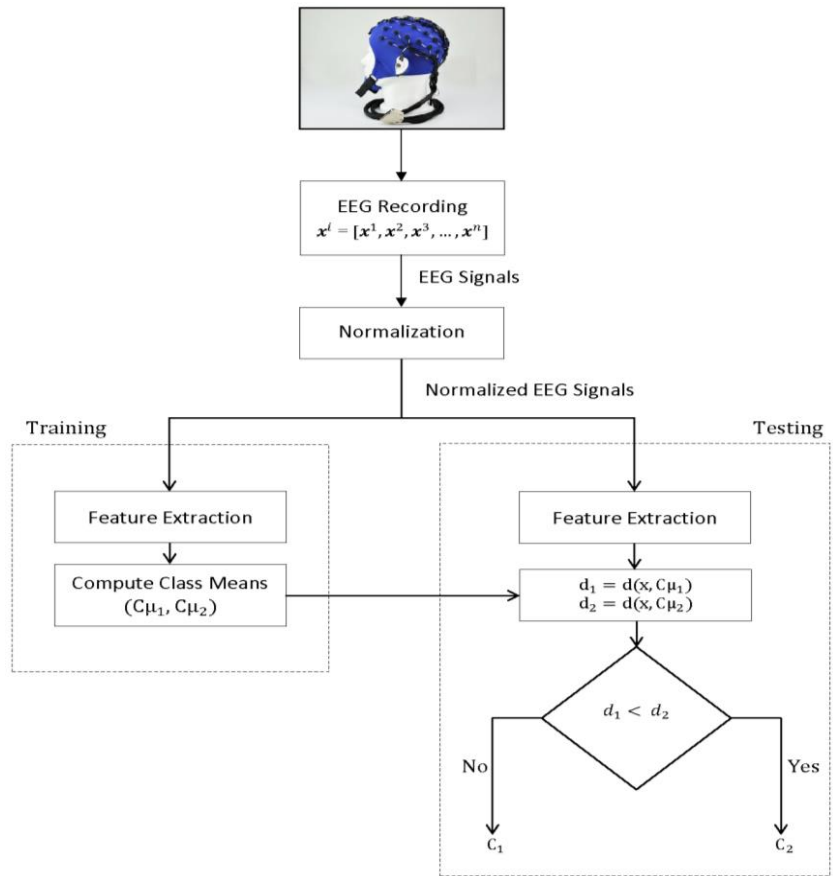

Fig. 1. Proposed methodology for feature extraction and classification of EEG Brain signals

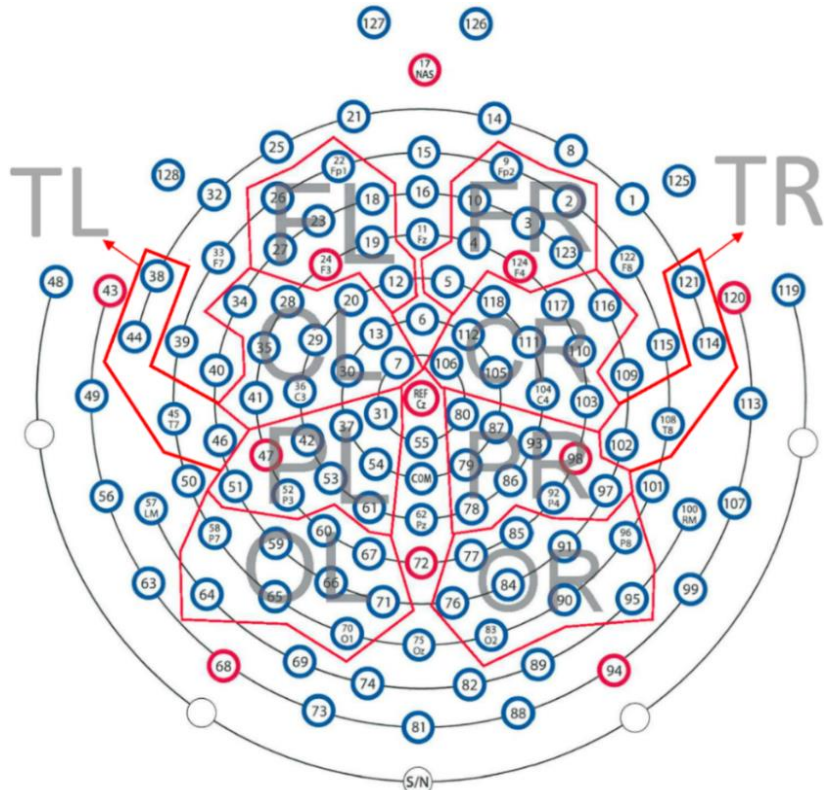

Fig. 2. Brain regions used for the analysis. TEMP: Temporal-right (TR) and temporal-left (TL); FRONT: frontal-right (FR) and frontal-left (FL); CENT: central-right (CR) and central-left (CL); PERI: parietal-right (PR) and parietal-left (PL); OCCIP: occipital-right (OR) and occipital-left (OL); ALL: All regions - 100 channels

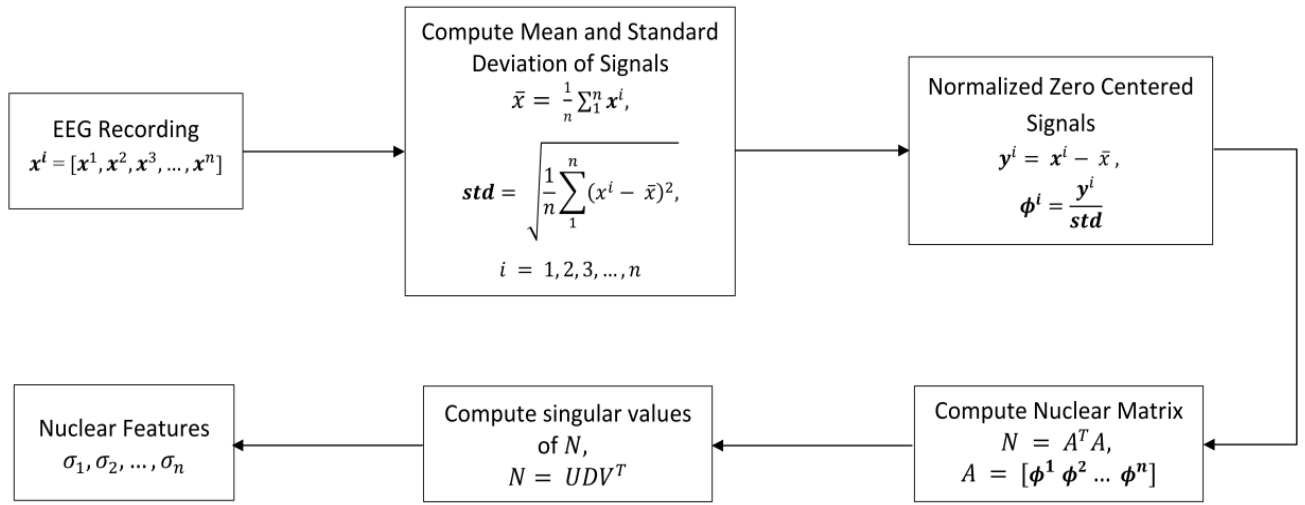

Fig. 3. Proposed feature extraction methodology

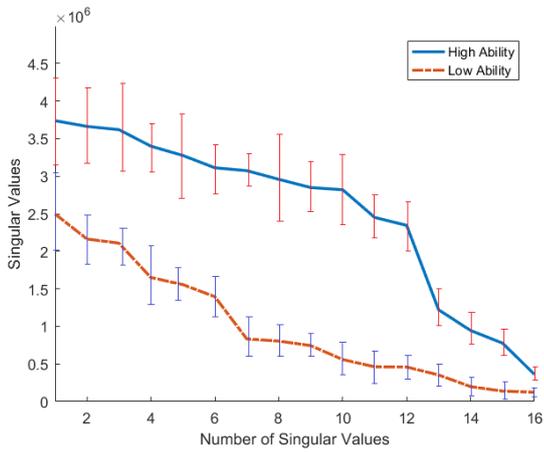

Fig. 4. Plot of singular values obtained from clean EEG signals captured from FRONT region of all subjects (FICD)

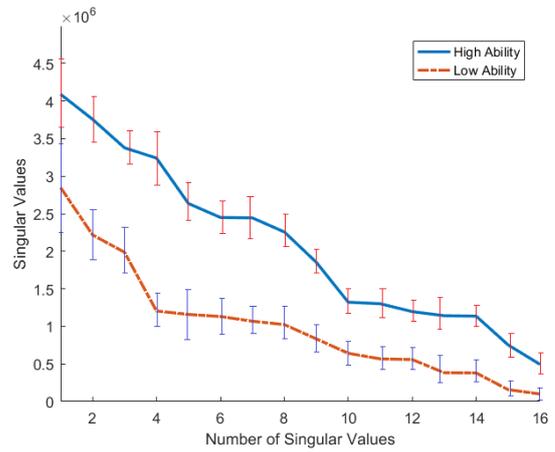

Fig. 5. Plot of singular values obtained from raw EEG signals captured from FRONT region of all subjects (FIRD)

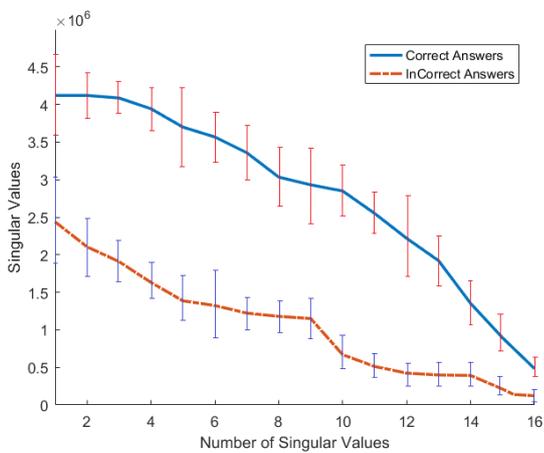

Fig. 6. Plot of singular values obtained from clean EEG signals captured from FRONT region of all subjects (MRT)

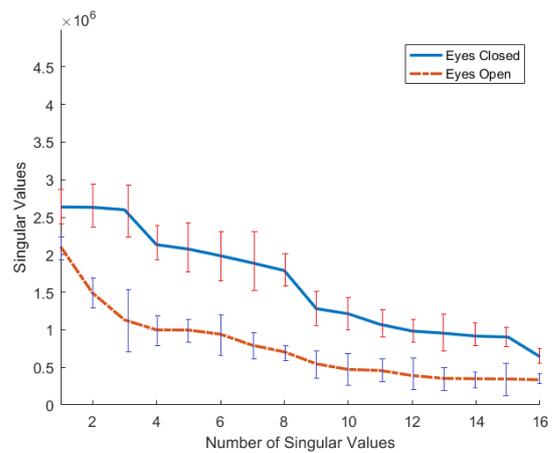

Fig. 7. Plot of singular values obtained from clean EEG signals captured from FRONT region of all subjects (EOEC)

To be precise, we denote the EEG signal corresponding to $i^{th}$ channel by $x^i$, where $x^i = [x_1^i(t_1), x_2^i(t_2), x_3^i(t_3), \ldots, x_d^i(t_d)]^T = [x_1^i, x_2^i, x_3^i, \ldots, x_d^i]$, to avoid clutter we drop the time stamps. Let $n$ be the number of electrodes placed on a particular brain region and $x^1, x^2, x^3, \ldots, x^n$ be the signals (channels) captured from them during brain activation corresponding to an event, i.e., these signals represent the brain state activated by an event and are treated as an instance. The mean and standard deviation of these signals are:

$$\overline{x} = \frac{1}{n}\sum_1^n x^i. \qquad (2)$$

$$std = \sqrt{\frac{1}{n}\sum_1^n (x^i - \overline{x})^2} \qquad (3)$$

First, we transform the signals so that they are zero-centered, i.e.,

$$y^i = x^i - \overline{x}, \; i = 1, 2, 3, \ldots, n \qquad (4)$$

After that each signal $y$ is divided component-wise by the standard deviation $std$ so that each component has unit variance:

$$\boldsymbol{\phi}^i = \frac{y^i}{std}, i = 1, 2, 3, \ldots, n \qquad (5)$$

Using the transformed signals, we define the following matrix:

$$N = A^T A, \qquad (6)$$

where $A = [\boldsymbol{\phi}^1 \; \boldsymbol{\phi}^2 \; \ldots \; \boldsymbol{\phi}^n]$. The size of $N$ is $n \times n$, and it represents a single event (e.g. high ability or low ability), we call it *Nuclear Matrix*. Using singular value decomposition (SVD), it is factorized as follows:

$$N = UDV^T \qquad (7)$$

where $D$ is diagonal and the diagonal entries are singular values $\sigma_i, i = 1, 2, 3, \ldots, n$. Fig. 4 - 7 shows the plots of singular values corresponding to two different events for visual oddball task, memory recall task and baseline task, respectively. These plots indicate that singular values clearly discriminate the two events, as such these can be used to differentiate the brain states corresponding to different events. Also, these plots show that the largest singular values are more discriminant. In view of this observation, we use the two largest singular values to represent the brain states stimulated by different events. Further to rule out the thought that the discrimination depicted in Fig. 4 - 7, is not only associated with two particular examples, we considered a number examples of two events for clean and raw data of visual oddball task, MRT, and EOEC task. Fig. 8 - 11 show the plots, where in each plot EEG brain signals, represented as two largest singular values, have been depicted as green crosses and red circles, which describe examples of two different events. The examples related to two different events cluster together in two distinct regions of the feature space, which can be separated by simple decision boundary. This observation leads to the conclusion that two largest singular values can discriminate well EEG brain signals corresponding to different events and can be used as features to represent the events, we call them as *nuclear features*. The histogram of first feature from HA and LA Class (FICD) is shown in Fig. 12.

Further, to have an insight into nuclear features, consider:

$$C = AA^T \qquad (8)$$

which is a covariance matrix of size $d \times d$ ($d >> n$), as $d$ is the number of samples in an EEG signal and $n$ is the number of signals (channels) recorded from a brain region) and gives the correlation structure among different EEG signals corresponding to an event. Its eigenvalues $\lambda_i, i = 1, 2, 3, \ldots, n$ represent the variances of the samples of EEG signals along principle directions, specified by the corresponding eigenvectors $\boldsymbol{u}_i, i = 1, 2, 3, \ldots, n$ of $C$ [39]. As $\lambda$ is an eigenvalue of $C$ corresponding to eigenvector $\boldsymbol{u}$, so

$$C\boldsymbol{u} = \lambda\boldsymbol{u} \qquad (9)$$

or it can be written as follows:

$$AA^T\boldsymbol{u} = \lambda\boldsymbol{u} \qquad (10)$$

Multiplying both sides on the left with $A^T$ yields:

$$A^T AA^T\boldsymbol{u} = \lambda A^T\boldsymbol{u} \qquad (11)$$

As $N = A^T A$, in view of this Eq. 11 becomes:

$$N(A^T\boldsymbol{u}) = \lambda(A^T\boldsymbol{u}) \qquad (12)$$

This indicates that $\lambda_i, i = 1, 2, 3, \ldots, n$ are also eigenvalues of $N$ corresponding to eigenvectors $A^T\boldsymbol{u}$, i.e., $\lambda_i$'s represent the variances of samples of EEG signals along the principle directions specifies by $A^T\boldsymbol{u}$'s. As $\lambda = \sigma^2$ [39], so the nuclear features represent variances of the samples of EEG signals corresponding to an event along principle directions.

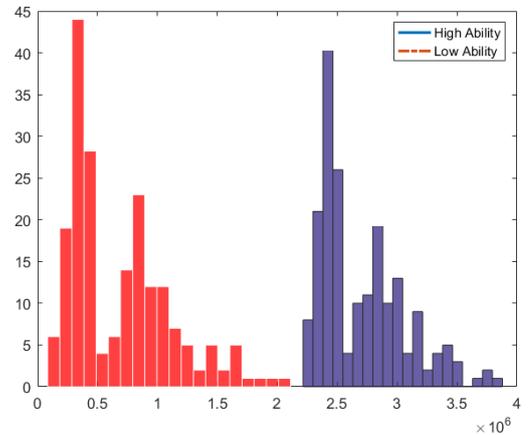

Fig. 12. Histogram of first feature from HA and LA Class (FICD)

*3.2. Class Means based Minimum Distance Classifier (CMMDC)*

From Fig. 8 - 11, it is observed that data belonging to each class is clustered around the mean of its own class. The plots show that the examples related to two different events have high interclass variation and are clustered in separate regions of the feature space, which can be separated by simple decision boundary. By keeping this observation, we decided to use simple and efficient minimum distance classifier based on class means to classify the nuclear features of two classes, i.e., LA and HA, correct and incorrect answers, and EO and EC.

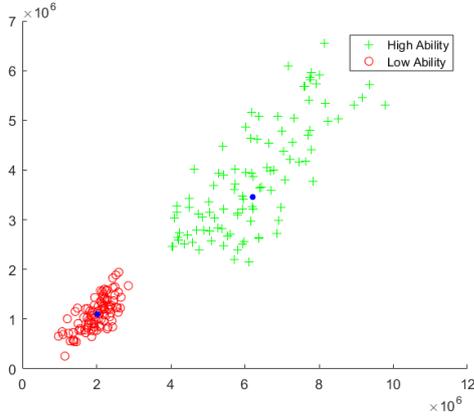

Fig. 8. Plot of examples of two groups, each subject in the groups is represented using two dominant nuclear features extracted from *clean EEG signal* from FRONT region (FICD).

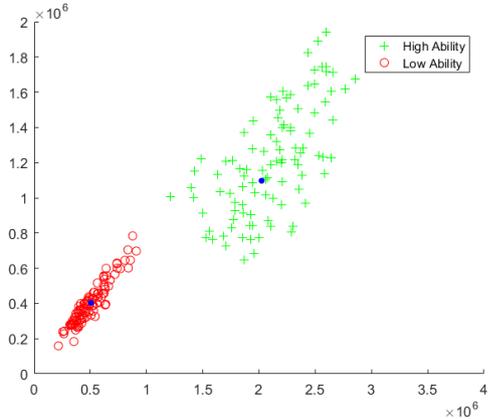

Fig. 9. Plot of examples of two groups, each subject in the groups is represented using two dominant nuclear features extracted from *raw EEG signal* from FRONT region (FIRD).

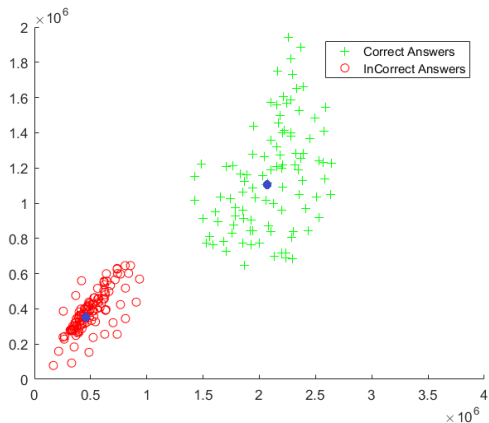

Fig. 10. Plot of examples of two groups, each subject in the groups is represented using two dominant nuclear features extracted from *clean EEG signal* from FRONT region (MRT).

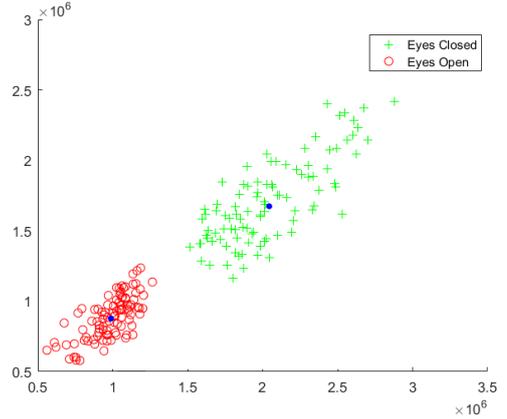

Fig. 11. Plot of examples of two groups, each subject in the groups is represented using two dominant nuclear features extracted from *clean EEG signal* from FRONT region (EOEC).

The CMMDC assigns a sample to the class for which distance between the sample and its mean is minimum. The distance as a measure of similarity indicates that similarity is maximum if distance is minimum. First we calculate the class means, i.e., $C\mu_1$ and $C\mu_2$ of both classes ($C_1$ and $C_2$) in each dataset. Then we calculate the distance between each sample $(x)$, and mean of each class as follows:

$$d_1 = d(x, C\mu_1) \quad (13)$$
$$d_2 = d(x, C\mu_2) \quad (14)$$

If $d_1 < d_2$ then $x \in C_2$ else $x \in C_1$.

Classification results on four datasets i.e., $FICD, FIRD, MRT,$ and $EOEC$ with two different brain regions, i.e., *FRONT* and *PERI* are shown in Tables 1 and 2.

In addition to CMMDC, we also used support vector machines (SVM) classifier with linear kernel for comparison.

To evaluate the classification system, we used ten-fold cross-validation method to evaluate the system performance under different dataset variations. The dataset was divided into ten different folds. In every turn, each data fold was taken out and other remaining nine folds were utilized to tune and train the system. After tuning and training the system, the held-out data fold was utilized as an independent set to evaluate the system performance. This procedure was repeated for each data fold and average performance values were computed. The main benefit of this method was that the proposed system was evaluated under different variations of dataset samples.

## 4. Experimental Results and Discussion

In this section, firstly, we present the evaluation protocol for the proposed scheme, then the results with nuclear features have been presented and discussed.

*4.1. Evaluation Protocol*

To evaluate the system performance, we employed the commonly used measures (specificity, accuracy and sensitivity) which are given below:

$$Sensitivity = \frac{TP}{TP+FN} \times 100 \quad (15)$$

$$Accuracy = \frac{TP+TN}{TP+FN+TN+FP} \times 100 \quad (16)$$

$$Specificity = \frac{TN}{TN+FP} \times 100 \quad (17)$$

where

*TP*: the number of true positives e.g. the number of subjects belonging to LA predicted as belong to LA,
*TN*: the number of true negatives e.g. the number of subjects belonging to HA predicted as belong to HA,
*FP*: the number of false positives e.g. the number of subjects belonging to HA predicted as belong to LA,
*FN*: the number of false negatives e.g. the number of subjects belonging to LA predicted as belong to HA.

We also used area under ROC curve as a performance measure.

In fluid intelligence prediction level datasets, i.e., $FICD$ and $FIRD$, the total numbers of trials for HA and LA groups were 551, and 482, respectively, and the length of channels in each trial was 150 samples. In MRT, the total number of correct and incorrect answers was 537 and 143, respectively. In EOEC dataset, baseline task was performed by each subject before the memory recall task for the duration of five (05) minutes.

*4.2. Performance with Nuclear Features*

To analyze the proposed system performance, we considered six brain regions, as shown in Fig. 2 and the clean and raw EEG signals captured from these regions. To assess the performance, we extracted nuclear features using 8, 16, 20, 18, 18 and 100 channels captured from *TEMP, FRONT, CENT, PERI, OCCIP* and *ALL* brain regions, respectively. After extracting the nuclear features from training data, we computed class means, and classified the test data using *CMMDC*, the results are shown in Fig. 13 - 16. The different regions lead to different results. The best performance in assessing the fluid intelligence level, memory recall task and baseline task is given by two regions out of six, i.e., $FRONT$ and $PERI$ regions; the accuracy for other regions is below 97%. These results show the dominance of $FRONT$ and $PERI$ regions, which gave best results, i.e., 100% and 99% accuracies, with two or three nuclear features. The detailed results on the four datasets, i.e., $FICD, FIRD, MRT, and EOEC$ for two brain regions, i.e., $FRONT$ and $PERI$ are shown in Table 1. This discussion indicates that nuclear features extracted from two regions ($FRONT$ and $PERI$) are discriminative and lead to the best results in classifying the subjects based on their fluid intelligence level, i.e., LA or HA, memory recall task and baseline task.

The results given in Table 1 for FICD and FIRD show that raw data gives results equivalent to that from clean data, in case PERI region FIRD results in relatively better performance than FICD. For PERI region, FIRD gives 99.5% accuracy and 99.8% sensitivity, whereas 99% accuracy and 99.4% sensitivity are obtained from FICD. The reason that FICD gives slightly less accuracy and sensitivity than FIRD is that some information is lost during cleaning the EEG signals. It follows from this discussion that nuclear features are robust in representing the raw data.

To validate the usefulness of CMMDC, we further employed SVM classifier with linear kernel; results with SVM classifier are shown in Table 2; it also gave 100% and 99% accuracies using nuclear features extracted from $FRONT$ and $PERI$ region, respectively. The results are same as obtained by CMMDC but CMMDC is more efficient than SVM in computational time; also CMMDC is memory efficient, it does not need to keep the whole training data, only two class means are stored. It indicates that the CMMDC is more suitable for the classification of subjects based on their fluid intelligence level, memory recall task and baseline task.

Fig. 17 and 18 show the area under the receiver operating characteristic curve (AUC) for $FRONT$ region with datasets, i.e., $FICD\ and\ FIRD$.

*4.3. System Performance using Different Subjects for Training and Testing Data*

In our evaluation of proposed system so far, we utilized all dataset samples collected from all subjects. For this purpose, we divided the dataset samples into training and testing data utilizing ten-fold cross-validation to evaluate the system performance over various dataset variations. In order to validate that the proposed system does not depend on the subjects, we carried out the experiments by utilizing testing and training data from different subjects. For this purpose, we separate out the subjects for testing from whom the dataset samples was not utilized for training purpose. For training and testing, we utilized the 90% and 10% of the subject's data, respectively. In *FICD and FIRD,* we used the nuclear features found to be the best in previous sections. Therefore, we achieved the 100% accuracy through SVM classifier for $FRONT$ region. It showed that the system performance is not dependent on the subjects.

*4.4. Analysis of Brain Regions*

In this study, we decomposed the entire brain region into six regions as shown in Fig. 2 to find out which region plays dominant role in fluid intelligence level prediction. In previous studies [40, 41], it has been shown that different brain regions deal with different tasks like intelligence, attention and cognitive tasks. Structural and functional neuroimaging studies found that fronto-parietal network consisting of $FRONT$ and $PERI$ deals with intelligence. Aron et al. [40] reported that the fronto-parietal network handles cognitive functions related to intelligence, perception, short-term memory storage, and language. Our findings about which brain region is more effective in assessing the fluid intelligence level of individuals also lead to the same conclusion and further corroborate that fronto-parietal network is concerned with fluid intelligence level.

Table 1

CMMDC (Class Means based Minimum Distance Classifier) results for the prediction of fluid intelligence level (LA vs HA), *MRT* and *EOEC* task [*FICD*: Fluid Intelligence Clean Data, *FIRD*: Fluid Intelligence Raw Data, *MRT*: Memory Recall Task, *EOEC*: Eyes open / Eyes Closed Data, AUC: area under the curve, FRONT: frontal-left (FL) and frontal-right (FR), PERI: parietal-left (PL) and parietal-right (PR), *LA*: low ability, *HA*: high ability]

| Dataset | Brain Region | No. of Nuclear Features | No. of Channels | Accuracy | Sensitivity | Specificity | AUC |
|---|---|---|---|---|---|---|---|
| *FICD* | *FRONT* | 02 | 16 | 100 | 100 | 100 | 1 |
|  | *FRONT* | 03 | 16 | 100 | 100 | 100 | 1 |
|  | *PERI* | 02 | 18 | 99.7 | 99 | 99 | 0.99 |
|  | *PERI* | 03 | 18 | 99 | 99.4 | 99 | 0.989 |
| *FIRD* | *FRONT* | 02 | 16 | 100 | 100 | 100 | 1 |
|  | *FRONT* | 03 | 16 | 100 | 100 | 100 | 1 |
|  | *PERI* | 02 | 18 | 99.1 | 99.5 | 98.9 | 0.98 |
|  | *PERI* | 03 | 18 | 99.5 | 99.8 | 99 | 0.99 |
| *MRT* | *FRONT* | 02 | 16 | 100 | 100 | 100 | 1 |
|  | *FRONT* | 03 | 16 | 100 | 100 | 100 | 1 |
|  | *PERI* | 02 | 18 | 99.2 | 99 | 99 | 0.998 |
|  | *PERI* | 03 | 18 | 99.7 | 99.1 | 99.4 | 0.99 |
| *EOEC* | *FRONT* | 02 | 16 | 100 | 100 | 100 | 1 |
|  | *FRONT* | 03 | 16 | 100 | 100 | 100 | 1 |
|  | *PERI* | 02 | 18 | 99.3 | 99.3 | 98.8 | 0.99 |
|  | *PERI* | 03 | 18 | 99.5 | 99.1 | 98.6 | 0.989 |

Table 2.

SVM (Support Vector Machines) classifier results for the prediction of fluid intelligence level (LA vs HA), MRT and EOEC task [*FICD*: Fluid Intelligence Clean Data, *FIRD*: Fluid Intelligence Raw Data, *MRT*: Memory Recall Task, *EOEC*: Eyes open / Eyes Closed Data, AUC: area under the curve, FRONT: frontal-left (FL) and frontal-right (FR), PERI: parietal-left (PL) and parietal-right (PR), *LA*: low ability, *HA*: high ability]

| Dataset | Brain Region | No. of Nuclear Features | No. of Channels | Accuracy | Sensitivity | Specificity | AUC |
|---|---|---|---|---|---|---|---|
| *FICD* | *FRONT* | 02 | 16 | 100 | 100 | 100 | 1 |
|  | *FRONT* | 03 | 16 | 100 | 100 | 100 | 1 |
|  | *PERI* | 02 | 18 | 99.4 | 99.6 | 98.2 | 0.99 |
|  | *PERI* | 03 | 18 | 99.1 | 99.3 | 98 | 0.987 |
| *FIRD* | *FRONT* | 02 | 16 | 100 | 100 | 100 | 1 |
|  | *FRONT* | 03 | 16 | 100 | 100 | 100 | 1 |
|  | *PERI* | 02 | 18 | 99 | 99.5 | 98.1 | 0.99 |
|  | *PERI* | 03 | 18 | 99.2 | 99.4 | 98.9 | 0.99 |
| *MRT* | *FRONT* | 02 | 16 | 100 | 100 | 100 | 1 |
|  | *FRONT* | 03 | 16 | 100 | 100 | 100 | 1 |
|  | *PERI* | 02 | 18 | 99 | 99.3 | 98.9 | 0.99 |
|  | *PERI* | 03 | 18 | 99.1 | 99.4 | 98.5 | 0.987 |
| *EOEC* | *FRONT* | 02 | 16 | 100 | 100 | 100 | 1 |
|  | *FRONT* | 03 | 16 | 100 | 100 | 100 | 1 |
|  | *PERI* | 02 | 18 | 99 | 99.4 | 98.4 | 0.989 |
|  | *PERI* | 03 | 18 | 99.3 | 99.4 | 98.4 | 0.99 |

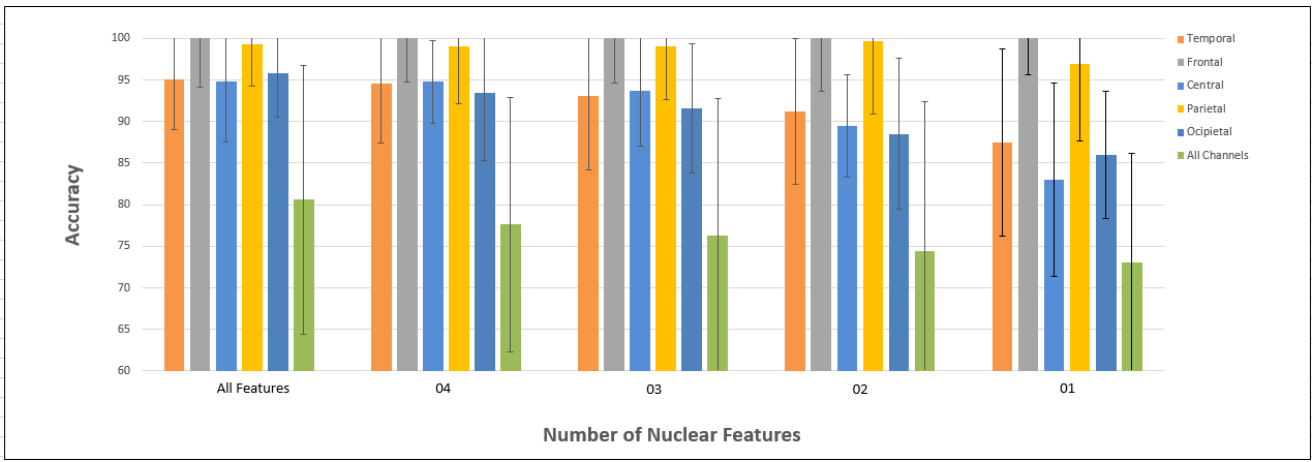

Fig 13. Classification results of nuclear features with CMMDC (Class Means based Minimum Distance Classifier) with fluid intelligence clean data (*FICD*)

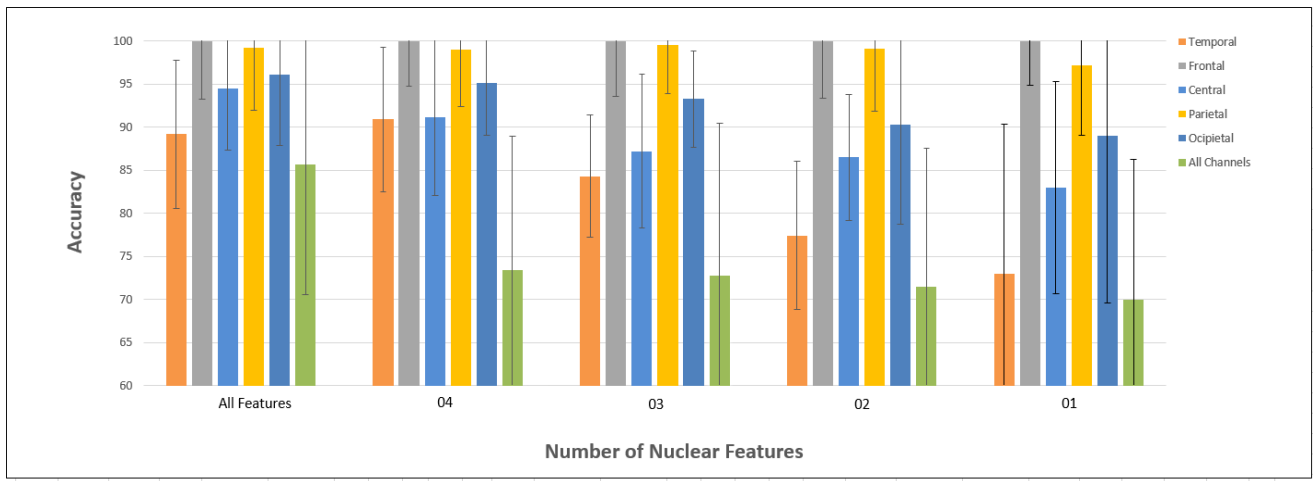

Fig 14. Classification results of nuclear features with CMMDC (Class Means based Minimum Distance Classifier) with fluid intelligence raw data (FIRD)

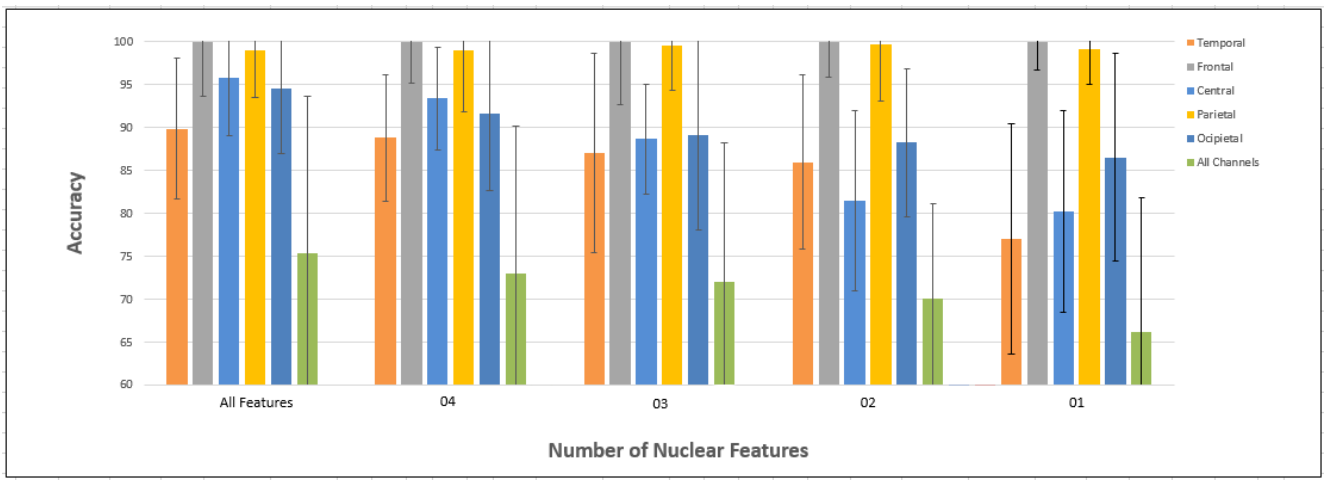

Fig 15. Classification results of nuclear features with CMMDC (Class Means based Minimum Distance Classifier) with Memory Recall (MRT) Data

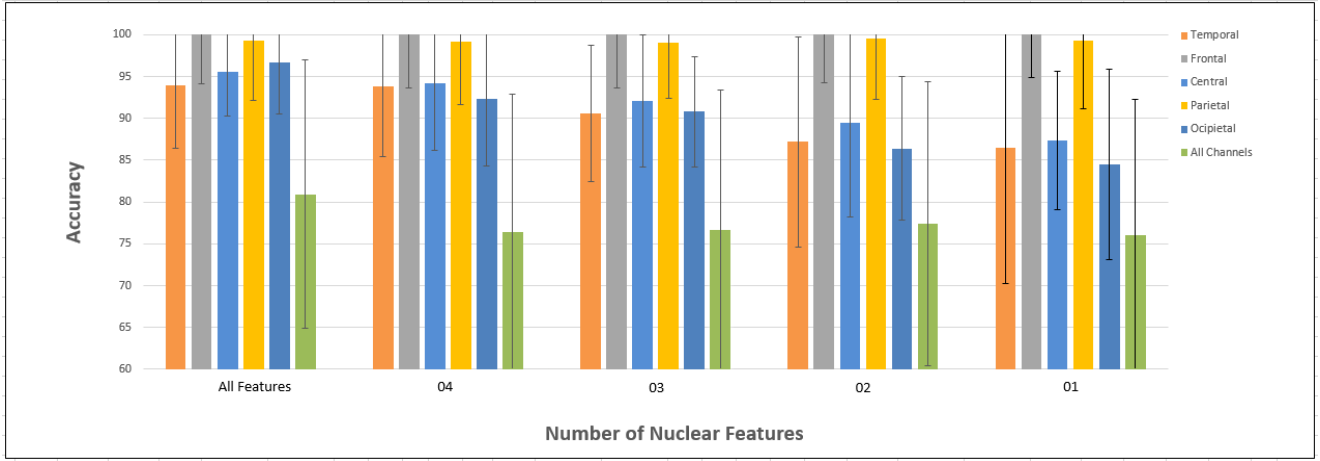

Fig 16. Classification results of nuclear features with CMMDC (Class Means based Minimum Distance Classifier) with Eyes open / Eyes Closed Data (EOEC)

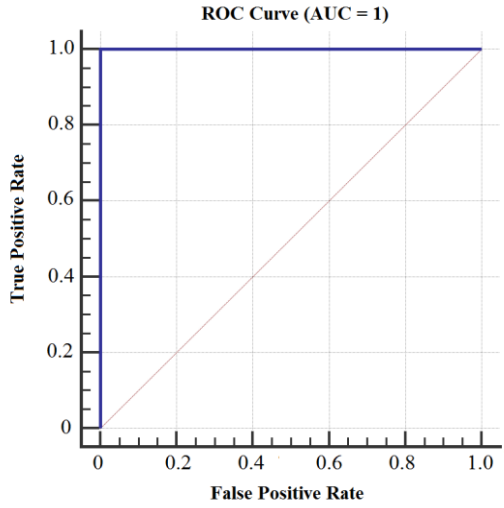

Fig. 17. Area under the receiver operating characteristic curve (AUC) for FRONT region with fluid intelligence (FICD)

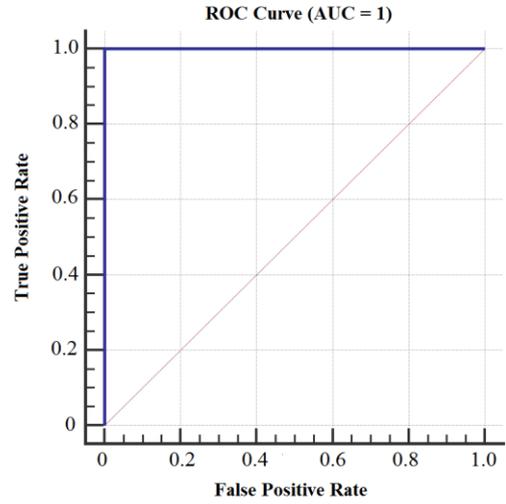

Fig. 18. Area under the receiver operating characteristic curve (AUC) for FRONT region with fluid intelligence (FIRD)

### 4.5. Interclass and Intraclass Variation Analysis

To show the significant difference between the nuclear features of classes belonging to all four datasets, we performed the interclass and intraclass variation analysis. For this purpose, scatter matrices [42] are used. We computed between class scatter matrix $(S_b)$ and within class scatter matrix $(S_w)$ of different classes in each dataset. After computing the $S_b$ and $S_w$, mixture scatter matrix $(S_m)$ is computed, i.e.,

$$S_m = S_b + S_w \quad (18)$$

Where $S_m$ shows the covariance matrix of the feature vector according to the global mean. Next, determinant and trace of $S_w$ and $S_m$ are computed. The criteria to check the significant difference between the nuclear features of two classes is that the trace and determinant of mixture scatter matrix $\{S_m\}$ should be greater than the trace and determinant of within class scatter matrix $\{S_w\}$, respectively. It is shown below;

$$J_1 = \frac{trace\{S_m\}}{trace\{S_w\}} \quad (19)$$

$$J_2 = \frac{det\{S_m\}}{det\{S_w\}} \quad (20)$$

It took larger values when samples in the L –dimensional space are well clustered around their mean, within each class and the clusters of the different classes are well separated. The results are shown in Table 3 after using Eqs. (19) and (20):

Table 3.

Inter-Class and Intra-Class Variation Analysis

| Dataset | J1 | J2 |
|---|---|---|
| FICD | 5.724 | 6.680 |
| FIRD | 11.0508 | 23.2570 |
| MRT | 10.42207 | 19.946 |
| EOEC | 6.6803 | 17.894 |

This indicates that criteria which define the condition, i.e., det $\{S_m\}$ > det $\{S_w\}$ and trace $\{S_m\}$ > trace $\{S_w\}$ holds. Therefore, analytical analysis shows that significant difference exists between the features of all classes in each dataset.

*4.6. Comparisons*

To check the effectiveness of our proposed system, we compared our experimental results with previous EEG research studies. Table 4 represents a detailed comparison of proposed methodology with previous research studies considering the performance of classifier, cognitive tasks, dataset, feature extraction methods and ML (machine learning) algorithm. The previous research studies as shown in Table 4 used various techniques for feature extraction such as autoregressive coefficients (AR), wavelet-transform, time domain and frequency domain based features for the classification of EEG brain signals, which are recorded during the cognitive task. Some of the studies used the non-linear classifiers as shown in Table 4. These classifiers were more time-consuming and complex to construct the classifier model. In [43], authors used small number of instances for the classification purpose that creates the overfitting problem. In our proposed methodology, a large number of samples or trials were used for each class. In the classification phase, we used ten (10) fold cross validation scheme as well as different subjects for training (90%) and testing data (10%).

The main benefit of this method was that the proposed system was evaluated under different variations of dataset samples in the training and testing phase [43]. From Table 4, we noted that the classification results of our study are better than the previous research studies that were using the different or same classifier with the similar nature of the cognitive task.

Table 4.

Comparison of proposed approach with existing techniques

| Sr | Ref | Subjects | Scalp Electrodes | Feature | Classifier | Accuracy | Cognitive task |
|---|---|---|---|---|---|---|---|
| 1 | Our Proposed Work | 34 | 16 | Nuclear features | MDC, SVM with linear Kernel | 100 % | Memory Recall, RAPM, Visual oddball cognitive and baseline tasks |
| 2 | [44] | 07 | 6 | Autoregressive coefficients | ELM, SVM and ANN | 53.98 to 56.07 | 05 tasks as mentioned in [45] |
| 3 | [46] | 07 | 6 | Frequency and time domain features | LDA and ANN | 87.35 to 91.17 | 05 tasks as mentioned in [45] |
| 4 | [47] | 03 | 8 | Power feature | SVM and ANN | 65.90 to 68.35 | 03 cognitive tasks (words generation, imagination of left and right hand movement) |
| 5 | [48] | 02 | 06 | Wavelet packet entropy | SVM | 87.5 to 93.0 | 05 tasks as mentioned in [45] |
| 6 | [49] | 04 | 06 | Discrete wavelet transform | ANN | 74.40 to 82.30 | 05 tasks as mentioned in [45] |
| 7 | [50] | 07 | 06 | Autoregressive coefficients and power of frequency bands | SVM | 70 | Three tasks (multiplication, baseline and mental letter) |
| 8 | [51] | 04 | 06 | db4 wavelet | K-NN | 81.48 to 89.58 | 05 tasks as mentioned in [45] |
| 9 | [52] | 07 | 06 | Immune feature | SVM | 85.40 to 97.5 | 05 tasks as mentioned in [45] |
| 10 | [53] | 04 | 06 | Wavelet packet | RBF network | 85.30 | 05 tasks as mentioned in [45] |
| 11 | [55] | 08 | 128 | Wavelet relative energy | SVM with RBF Kernel | 98.75% | RAPM, Visual oddball cognitive and baseline tasks |

*4.7. Limitations of the Study*

There are few limitations in this research study, which will be given consideration in future research. In this study, all the subjects were male. However, female subjects will also be considered in our future study and experiments to predict the memory recall ability and fluid intelligence level of both the genders. To investigate the diseases effects on memory recall ability and fluid intelligence level, some subjects will also be considered in future study that might have some medical problems. In this study, there were only thirty-four (34) subjects; therefore, number of subjects should be increased to validate that EEG brain signals would be enough to predict the cognitive performance. Moreover, this research study examined the association of EEG brain signals with learning ability and memory for young subjects only. In addition, the learning materials used in this research study were related to anatomy material and physiology; so, the results cannot be generalized to relate with learning capability of many other types of academic learning materials or memory recall capability.

## 5. Conclusion

A method for the classification of raw EEG brain signals has been proposed, which is based on an efficient approach for the extraction of discriminative nuclear features and, simple and memory-efficient CMMDC that requires only the mean feature vector of each class. The method is robust and computationally efficient for raw as well as clean data. In spite of having feature space with small dimension, it gives better results as compared to other techniques. We validated the robustness and effectiveness of the method on fluid intelligence level prediction problem, i.e., LA or HA, MRT, and EOEC datasets. In this case, nuclear features were extracted from clean and raw EEG datasets captured from six different brain regions with 8, 16, 20, 18, 18 and 100 channels. The nuclear features from *FRONT* and *PERI* regions resulted in 100% and 99% prediction accuracies, respectively. The analysis showed that there is a clear difference between nuclear features corresponding to subjects belonging to different groups (classes). The discriminative potential of the features established the superiority of the proposed system. The proposed method is useful for different classification problems based on EEG brain signals.

**Conflicts of interest:** The authors have no conflicts of interest to declare.